\documentclass{article}
\usepackage{ijcai22}

\usepackage{times}
\usepackage{soul}
\usepackage{url}
\usepackage[hidelinks]{hyperref}
\usepackage[utf8]{inputenc}
\usepackage[small]{caption}
\usepackage{graphicx}
\usepackage{amsmath}
\usepackage{amsthm}
\usepackage{booktabs}
\usepackage{algorithm}
\usepackage{algorithmic}
\usepackage{algorithm}
\usepackage{algorithmic}

\usepackage{amsmath}
\usepackage{array}
\usepackage{bm}
\usepackage{enumerate}

\usepackage{bbm}
\usepackage{amssymb}
\usepackage{subfig}

\usepackage{cleveref}
\usepackage{booktabs}       

\usepackage{comment}

\usepackage{xcolor}

\title{Forming Effective Human-AI Teams: Building Machine Learning Models that Complement the Capabilities of Multiple Experts}

\author{
Patrick Hemmer$^1$\footnote{These authors contributed equally.}\and 
Sebastian Schellhammer$^{1,2*}$\and
Michael Vössing$^1$\and\\
Johannes Jakubik$^1$\And
Gerhard Satzger$^1$\\
\affiliations
$^1$Karlsruhe Institute of Technology\\
$^2$GESIS - Leibniz Institute for the Social Sciences
\emails
\{patrick.hemmer, michael.voessing, johannes.jakubik, gerhard.satzger\}@kit.edu,
sebastian.schellhammer@gesis.org
}

\begin{document}

\maketitle

\begin{abstract}
Machine learning (ML) models are increasingly being used in application domains that often involve working together with human experts. In this context, it can be advantageous to defer certain instances to a single human expert when they are difficult to predict for the ML model. While previous work has focused on scenarios with one distinct human expert, in many real-world situations several human experts with varying capabilities may be available. In this work, we propose an approach that trains a classification model to complement the capabilities of multiple human experts. By jointly training the classifier together with an allocation system, the classifier learns to accurately predict those instances that are difficult for the human experts, while the allocation system learns to pass each instance to the most suitable team member---either the classifier or one of the human experts. We evaluate our proposed approach in multiple experiments on public datasets with ``synthetic'' experts and a real-world medical dataset annotated by multiple radiologists. Our approach outperforms prior work and is more accurate than the best human expert or a classifier. Furthermore, it is flexibly adaptable to teams of varying sizes and different levels of expert diversity.
\end{abstract}

\section{Introduction}

Over the last years, the performance of machine learning (ML) models has become comparable to the one of human experts in a rising number of application domains \cite{esteva2017dermatologist}. 
For example, in medicine, several studies have demonstrated that ML models can surpass the performance of radiologists for the diagnosis of pneumonia in chest X-rays \cite{Chexpert} or achieve similar performance for the detection of diabetic retinopathy \cite{Gulshan2019PerformanceOA}. Despite these competitive results, individual experts are often still more accurate than ML algorithms in distinct areas of the input space \cite{TheAlgorithmicAutomationProblem}, e.g., due to limited model capacity, limited training data, or the existence of side information that is not accessible to the ML model \cite{LearningToComplementHumans}. In this context, prior work has investigated whether these different capabilities can be leveraged to enable ML models to complement the weaknesses of a single human expert by learning to predict only a fraction of the instances and pass the remaining ones to the domain expert \cite{Bansal2021,cuha2021,madras2018predict,ConsistentEstimators,TheAlgorithmicAutomationProblem,LearningToComplementHumans}. 
Even though in many real-world applications multiple human experts are available for load sharing and covering different instances \cite{pmlr-v81-chouldechova18a,gronsund2020augmenting}, 
the majority of these studies focus on settings in which ML models take the decision themselves for a given instance or leave it to a single human expert. Consequently, they do not leverage potential disagreement between individual human experts due to diverse capabilities. Thus, from a team perspective, the potential to improve the overall performance remains unused. Differences in the predictions of domain experts can, e.g., stem from different levels of knowledge or personal biases \cite{Lampert2016AnES}. Additionally, the rapid developments in highly specialized fields, e.g., medicine, are making it increasingly difficult for domain experts to acquire all knowledge within a single specialization domain. This creates the need to identify the appropriate human expert to assign particular instances to when they are difficult to predict for the ML model. We, therefore, propose an approach that jointly (1) trains a classifier to specifically complement the individual weaknesses of multiple human experts and (2) trains an allocation system to assign each instance either to the classifier or one specific human expert. The setting with several available human experts and a classifier is referred to as a team. The team member to whom an instance is assigned makes then a prediction on behalf of the entire team. Overall, our work makes the following contributions:

First, we present a novel approach that optimizes the performance of human-AI teams by jointly training a classifier together with an allocation system under consideration of the individual capabilities of available human experts. While the classifier learns to accurately predict the areas of the input space in which human experts are less competent, the allocation system learns to assign each instance to the best-suited team member. 

Second, we demonstrate that our approach improves the state-of-the-art by conducting experiments on two datasets with synthetically generated human experts---a dataset consisting of tweets containing hate speech and offensive language \cite{OffensiveLanguageDataset} and the popular CIFAR-100 dataset. We show that our approach can not only deal with different team sizes but also better leverages high human expert diversity in terms of capabilities than prior work \cite{TowardsUnbiasedAndAccurateDeferral}.
Moreover, we visually illustrate that the classifier in our approach makes correct predictions for the instances where human experts are less accurate and that the allocation system learns to assign exactly these instances to the classifier.  

Third, we demonstrate the real-world applicability of our approach with an example from the medical domain. The Chest X-ray dataset \cite{NIHDataset,chestxray8} of the NIH Clinical Center provides a suitable setting as the annotations do not only include radiologist-adjudicated labels, which serve as a high-quality ``gold standard'', but also the annotations on an individual radiologist level. In this context, prior work has conducted experimental evaluations on datasets that either provide only a distribution of human expert labels, e.g., CIFAR-10H \cite{Peterson2019HumanUM} or labels that can not be traced back to individual human experts, e.g., Galaxy Zoo dataset \cite{Lintott2008GalaxyZM}. 

We provide our code and the Appendix of this paper at \url{https://github.com/ptrckhmmr/human-ai-teams}.

\section{Related Work}

In recent years, there has been growing interest in optimizing the performance of human-AI teams---teams consisting of both a human expert and a classifier \cite{hemmer2021human}---by allocating a subset of the instances to a single human expert \cite{cuha2021,TowardsUnbiasedAndAccurateDeferral,madras2018predict,ConsistentEstimators,okati2021differentiable,TheAlgorithmicAutomationProblem,LearningToComplementHumans}. 
In this context, \citeauthor{TheAlgorithmicAutomationProblem} \shortcite{TheAlgorithmicAutomationProblem} separately train a classifier and a model for predicting whether a human expert errs. By deferring based on which of the two models has higher uncertainty, superior team performance can be achieved that surpasses both full automation (i.e., the classifier predicts all instances) and sole human effort (i.e., the human expert predicts all instances). While the authors train the classifier in isolation to optimize its performance, different studies emphasize that the classifier could also be trained to specifically complement the human expert's strengths and weaknesses. Several approaches propose to jointly train a classifier and a deferral system 
\cite{madras2018predict,LearningToComplementHumans}. Other approaches utilize objective functions with theoretical guarantees \cite{cuha2021,ConsistentEstimators}. \citeauthor{cuha2021} \shortcite{cuha2021} focus on support vector machines and maximize submodular functions to optimize team performance. \citeauthor{ConsistentEstimators} \shortcite{ConsistentEstimators} optimize a consistent surrogate loss derived from cost-sensitive learning. 
Further related work has investigated optimal triage policies \cite{okati2021differentiable}, user-initiated defer options \cite{Bansal2021}, or learning complementarity from bandit feedback \cite{gao2021human}. In general, these approaches focus on combining a classifier with a single human expert. However, in reality, more than one human expert is often available \cite{pmlr-v81-chouldechova18a,gronsund2020augmenting}. 
Hereby, it can be of particular interest in resource-scarce domains to consult exactly the right expert to avoid potentially costly mistakes. Our approach extends the idea that the classifier should complement the strengths and weaknesses of a single human expert to a setting with multiple human experts. The system allocates instances to one expert of a team consisting of multiple human experts and a complementary classifier. Concurrent work proposes a multi-label classification approach in which a classifier either makes a prediction or defers the instance to one of multiple human experts \cite{TowardsUnbiasedAndAccurateDeferral}. However, the approach does not leverage the human experts' disagreement to enhance the overall team performance but solely relies on identifying the best overall team member instead.

\section{Problem Formulation}

In this section, we introduce the problem of optimizing the performance of human-AI teams on classification tasks. Given a training example \(\bm{x} \in \mathcal{X}\), we learn to predict its ground truth label \(y \in \mathcal{Y} = \{1, \ldots, k\}\) where \(k\) denotes the number of classes. In addition to \(y\), we assume to have access to the predictions \(\bm{h} \in \mathcal{H}\) of \(m\) human experts for all training instances to implicitly learn their strengths and weaknesses. Each \(\bm{h}\) is a \(m\)-dimensional vector with \(\bm{h} = \begin{bmatrix} h_{1}, \ldots, h_{m} \end{bmatrix}\) with \(h_j \in \mathcal{Y} \,\, \forall j \in \{1, \dots, m\}\). We combine this data to the training dataset \(\mathcal{D}=\left\{(\bm{x},y, \bm{h})\right\}_{1}^N\) \(\sim P\), with \(N\) denoting the number of data instances, to jointly train an allocation system \(A: \mathcal{X} \rightarrow \mathbb{R}^{m+1}\) and a classifier \(F: \mathcal{X} \rightarrow \mathbb{R}^k\). Here, \(P\) is an (unknown) joint distribution, \(m+1\) denotes the total size of the human-AI team, and the outputs of \(A\) and \(F\) refer to the penultimate layer of the respective models. The classifier predicts \(\hat{y}_{clf} = \arg \max F(\bm{x})\) and the allocation system indicates the competence of each team member including the classifier with \(\bm{a} = A(\bm{x})\) for an instance \(\bm{x}\). Combining the individual predictions of all team members \(\hat{y} = [h_1, \ldots, h_m, \hat{y}_{clf}]\) and the index of the most competent team member \(j_{max} = \arg\max \bm{a}\) results in the team prediction \(\hat{y}_{team} = \hat{y}_{j_{max}}\). The goal is then to minimize the team loss
\begin{equation}
\mathcal{L}_{team}(F, A, \bm{x}, y, \bm{h}) = \mathbf{E}_{(\bm{x},y, \bm{h}) \sim P\ }\big[l(y, \hat{y}_{team}) \big].
\label{eq:expected_loss}
\end{equation}
To minimize the loss, the classifier needs to be accurate on those instances where the human experts are less accurate. The allocation system's function is to assign each instance to the team member with the highest probability of a correct prediction to answer on behalf of the team.

\section{Approach} 
In this section, we present our approach that jointly trains a classifier to complement the capabilities of multiple human experts together with an allocation system that assigns instances to either one of the human experts or the classifier. We first introduce the surrogate loss used to train both components followed by formalizing the approach as an algorithm.

Our approach draws upon the mixture of experts (MoE) framework that combines multiple classifiers with a gating network and trains them simultaneously \cite{Jacobs1991}. The gating network combines the outputs of the classifiers and partitions the input space such that each classifier is trained to predict the outputs for one partition. However, contrary to the MoE framework, we combine the outputs of one classifier \(F\) with the fixed predictions of the human experts. Besides using the allocation system \(A\) as a gating network, we further employ it to allocate instances among team members. 
Given the output of the allocation system \(\bm{a} = A(\bm{x})\), we obtain the probability \(w_j\) that team member \(j\) produces the correct output for an instance \(\bm{x}\) using
\begin{align}
w_j &= \frac{e^{a_j}}{\sum_{l=1}^{m+1}e^{a_l}}  &j = 1,\ldots,m+1\,.
\label{eq:approach_def_softmax}
\end{align}
The classifiers' conditional probability distribution over the ground truth labels \(\mathcal{Y}\) is given by 
\begin{align}
c_i &= \frac{e^{z_i}}{\sum_{l=1}^{k}e^{z_l}}  &i = 1,\ldots,k,
\label{eq:approach_clf_softmax}
\end{align}
where \(\bm{z} = F(\bm{x})\) is the output of the classifier. 
We then combine the conditional probabilities \(\bm{c}\) with the one-hot encoded predictions of the human experts to
\begin{equation}
   \bm{T}=\begin{bmatrix}\bm{h}_1,\ldots, \bm{h}_m, \bm{c}\end{bmatrix},
\end{equation}
with $\bm{h_j} \in \{0,1\}^{k}$. Using the predictions of all team members \(\bm{T}\) and the probabilities \(\bm{w}\), we derive the conditional probabilities on a team level
\begin{align}
P_{team}(Y=i|\bm{x}) &= \sum_{j=1}^{m+1} w_j \, T_{i,j}  &i = 1,\ldots,k \, .
\label{eq:approach_pteam}
\end{align}
Finally, using the team prediction and the one-hot encoded ground truth label \(\bm{y}\), we calculate the cross-entropy loss 
\begin{equation}
\mathcal{L}_{team}(F, A, \bm{x}, \bm{y}, \bm{h}) = - \sum_{i=1}^{k} y_{i}\,log(P_{team}(Y=i|\bm{x}))
\label{eq:approach_loss}
\end{equation}
for each instance to jointly train the allocation system and the classifier. Using the predictions of all team members weighted with the softmax probabilities of the allocation system in \Cref{eq:approach_pteam} is a soft relaxation of the team prediction used as input in \Cref{eq:expected_loss} to ensure differentiability. We formalize our approach in \Cref{alg:training_our_approach} and provide an intuition about the loss function's inner workings based on an analysis of the partial derivatives with respect to the classifier and allocation system in Appendix B. 
\begin{algorithm}
\caption{Training of our approach}
\begin{algorithmic}
\REQUIRE{\(\mathcal{D}=\left\{(\bm{x},\bm{y}, \bm{h}_{1},\ldots,\bm{h}_{m})\right\}_{1}^N\), the training data with \(\bm{y}\) and \(\bm{h}_{j}\) being one-hot encoded}
\REQUIRE{\(F\), the classifier with \(F: \mathcal{X} \rightarrow \mathbb{R}^{k}\)}
\REQUIRE{\(A\), the allocation system with \(A: \mathcal{X} \rightarrow \mathbb{R}^{m+1}\)}
\FOR{number of iterations} 
\STATE{sample a minibatch \(B\subseteq\{1,\ldots,N\}\)}
\FOR{\(i \in B\)}
\STATE{\(\bm{a} \gets A(\bm{x}^{(i)})\)}
\FOR{\(j = 1,\ldots,m+1\)}
\STATE{\(w_j \gets \frac{e^{a_j}}{\sum_{l=1}^{m+1}e^{a_l}}\)}
\ENDFOR
\STATE{\(\bm{z} \gets F(\bm{x}^{(i)})\)}
\FOR{\(l = 1,\ldots,k\)}
\STATE{\(c_l \gets \frac{e^{z_l}}{\sum_{o=1}^{k}e^{z_o}}\)}
\ENDFOR
\STATE{\(\bm{T} \gets \begin{bmatrix}\bm{h}^{(i)}_1,\ldots, \bm{h}^{(i)}_m, \bm{c}\end{bmatrix}\)}
\FOR{\(l = 1,\ldots,k\)}
\STATE{\(P_{team}(Y=l|\bm{x}^{(i)}) \gets \sum_{j=1}^{m+1} w_j \, T_{l,j}\)}
\ENDFOR
\STATE{\(\mathcal{L}^{(i)} = 
- \sum_{l=1}^{k} y^{(i)}_{l}\,log(P_{team}(Y=l|\bm{x}^{(i)}))\)}
\ENDFOR
\STATE{Backpropagate \(\mathcal{L}_{team} = \frac{1}{\lvert B\rvert}\sum_{i\in B}\mathcal{L}^{(i)}\)}
\ENDFOR
\end{algorithmic}
\label{alg:training_our_approach}
\end{algorithm}

\section{Experiments}

In this section, we evaluate the performance of our approach for different settings with synthetic and real-world human experts. First, we show that our approach is applicable to teams of varying sizes on a hate speech detection dataset from Twitter \cite{OffensiveLanguageDataset} and on the CIFAR-100 dataset \cite{krizhevsky2009learning}. Second, we investigate the influence of human experts' diversity with regard to their capabilities on the team performance using CIFAR-100 and visually illustrate the inner workings of our approach. The presence of 100 sub- and 20 superclasses in CIFAR-100 allows us to model human experts' capabilities with more granularity. Third, we assess its real-world applicability by presenting results on a chest X-ray dataset from the NIH Clinical Center \cite{NIHDataset,chestxray8} containing real radiologists' annotations. Additional experimental results and implementation details are provided in Appendix A. 

\paragraph{Baselines.} We compare the performance of our approach (\textit{Classifier \& Expert Team}) to multiple baselines. The first is the Joint Sparse Framework (\textit{JSF}) by \citeauthor{TowardsUnbiasedAndAccurateDeferral} \shortcite{TowardsUnbiasedAndAccurateDeferral}. In contrast to our approach, the authors pursue a \textit{multi-label} classification approach by training \(m + 1\) binary sigmoid classifiers and assigning each instance to the team member with the highest sigmoid weight. The second is a single classifier using a cross-entropy loss to predict all instances (\textit{One Classifier}) with the same architecture as the classifier in our approach. For the third baseline, we allocate instances randomly to one human expert from the team (\textit{Random Expert}). For the fourth baseline, we allocate all instances to the most accurate human expert in the team (\textit{Best Expert}).

Moreover, we report two additional baselines to investigate the contribution of the classifier and human experts to the team performance. Both baselines are simplified versions of the approach outlined in this work. For the first, we replace all \(m\) human experts with the same number of classifiers resulting in a \textit{Classifier Team} of size \(m+1\). These have the same architecture as the classifier in our approach and are jointly trained with the allocation system according to \Cref{alg:training_our_approach}. For the second, we omit the classifier resulting in an \textit{Expert Team} of size \(m\). The human experts are then used to train the allocation system using \Cref{alg:training_our_approach}.

\subsection{Hate Speech and Offensive Language Dataset}\label{sec:section_hate_speech}
We first conduct experiments on a Twitter dataset containing hate speech and offensive language \cite{OffensiveLanguageDataset}.

\paragraph{Data and Expert Generation.} The dataset consists of 24,783 tweets labeled as hate speech, offensive language, or neither. We follow the procedure for dataset preprocessing as described in \citeauthor{TowardsUnbiasedAndAccurateDeferral} \shortcite{TowardsUnbiasedAndAccurateDeferral}. We transform it into a binary classification task with \(1\) indicating hate speech or offensive language and \(0\) indicating neither of them. The tweets are additionally annotated with a second label indicating their dialect (i.e., African-American English (AAE) or non-African-American English (non-AAE)). We create \(m\) synthetic experts using this dialect label. For each expert, we sample the quantities \(p, q \sim \mathcal{U}(\cdot)\), with \(\mathcal{U}(\cdot)\) being the uniform distribution. The probability of correctly classifying non-AAE tweets is denoted by \(p\), whereas \(q\) represents the probability of correctly classifying AAE tweets. For \(\lfloor \frac{3m}{4}\rfloor\) experts, we define \(p \sim \mathcal{U}(0.6,1)\) and \(q \sim \mathcal{U}(0.6, p)\). For the remaining \(\lceil \frac{m}{4}\rceil\) experts, we set \(q \sim \mathcal{U}(0.6,1)\) and \(p \sim \mathcal{U}(0.6, q)\). As a result, \(\lfloor \frac{3m}{4}\rfloor\) experts are more accurate on non-AAE tweets and \(\lceil \frac{m}{4}\rceil\) experts are more accurate on AAE tweets.

\paragraph{Model.} We use pretrained GloVe word embeddings \cite{pennington2014glove} to generate 100-dimensional feature vectors. These features are used to train our approach and the algorithmic baselines. The classifier and allocation system are modeled by a neural network consisting of a single hidden layer with 50 units followed by a ReLU activation.

\paragraph{Experimental Setup.} To investigate the influence of the team size, we generate 2 to 20 synthetic human experts. The dataset is divided into a train, validation, and test split with 80\%, 10\%, and 10\% of the data. We train the models for 100 epochs using Adam optimizer with a learning rate of \(5 \cdot 10^{-3}\), a cosine annealing learning rate scheduler, and a batch size of 512. We apply early stopping on the validation split and report the results on the test split. We repeat the experiments 5 times with different seeds.
\begin{figure}[b]
    \centering
    \includegraphics[width=0.85\linewidth]{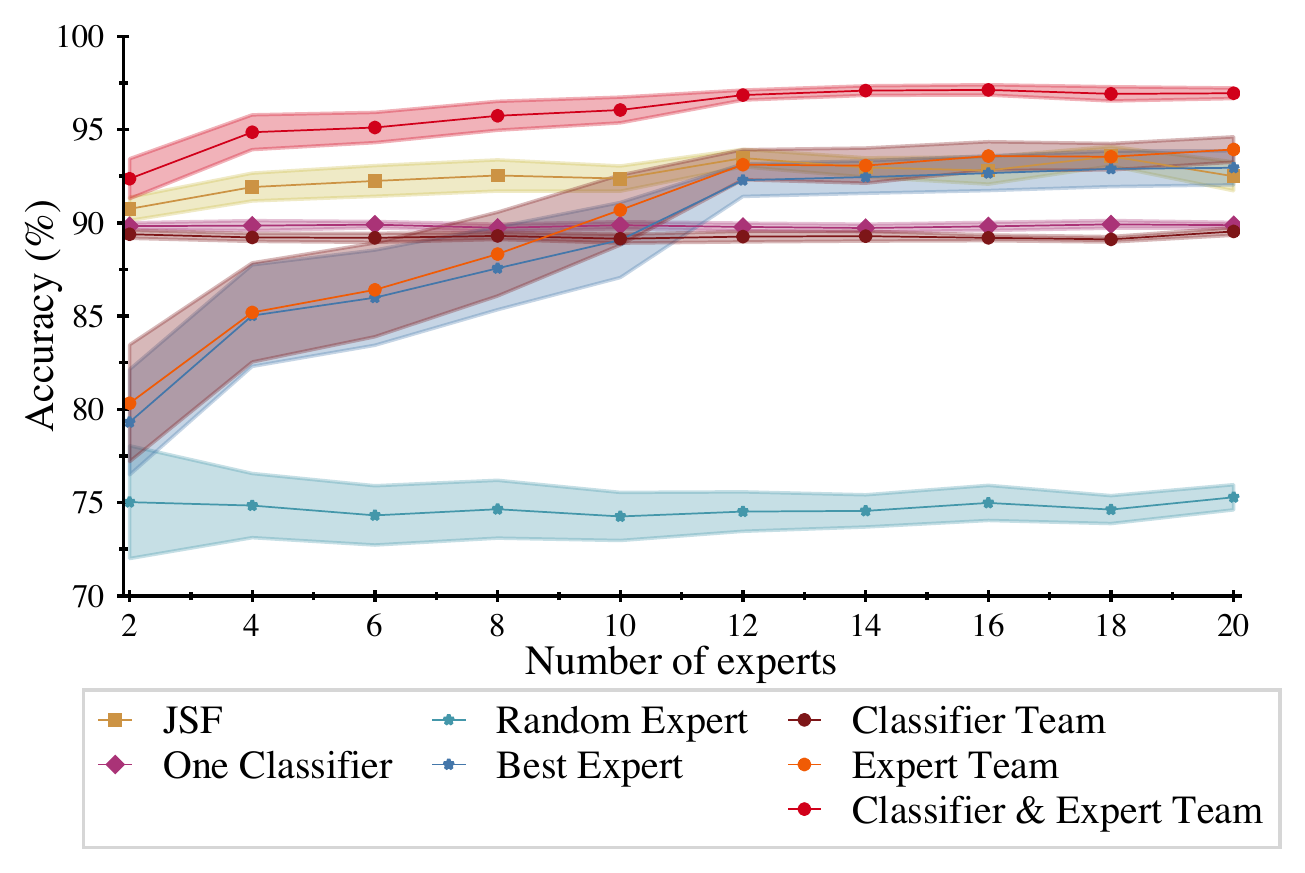}
     \caption{Team accuracy of our approach and the baselines with increasing team size on the Twitter dataset. Shaded regions display standard errors.}
    \label{fig:hate_speech_number_of_experts}
\end{figure}

\paragraph{Results.} We display the team accuracy of our approach and the respective baselines over the number of human experts in \Cref{fig:hate_speech_number_of_experts}. We see that all methods except the \textit{Random Expert} baseline drawing on human experts exhibit an increase in the overall performance with growing team size. This aligns with our expectation as with increasing team size more accurate human experts might enter the team and the covered feature space in which at least one team member is likely to perform an accurate prediction potentially increases. When taking a closer look, we find that our approach (\textit{Classifier \& Expert Team}) outperforms all baselines. The average improvement across all team sizes is 3.66\% over the \textit{JSF} baseline, 6.76\% over the \textit{One Classifier} baseline, 7.92\% over the \textit{Best Expert} baseline, 28.38\% over the \textit{Random Expert} baseline, and 7.44\% over the \textit{Classifier Team}. From the 6.96\% performance improvement of our approach over the \textit{Expert Team}, it becomes evident that the classifier in our approach learns to accurately predict the areas of the feature space that are not covered by human experts' competencies.

\subsection{CIFAR-100 Dataset}\label{sec:cifar-100}

The analysis in this section is threefold and uses the CIFAR-100 dataset \cite{krizhevsky2009learning}. First, we provide further experimental evidence on the performance of our approach with varying team sizes. Second, we investigate the influence of the diversity of human experts' capabilities on our approach. Third, we visualize how our approach distributes instances among classifier and human experts relative to each team member's respective competencies.

%
%

\begin{figure}[b]
    \centering 
    \includegraphics[width=0.85\linewidth]{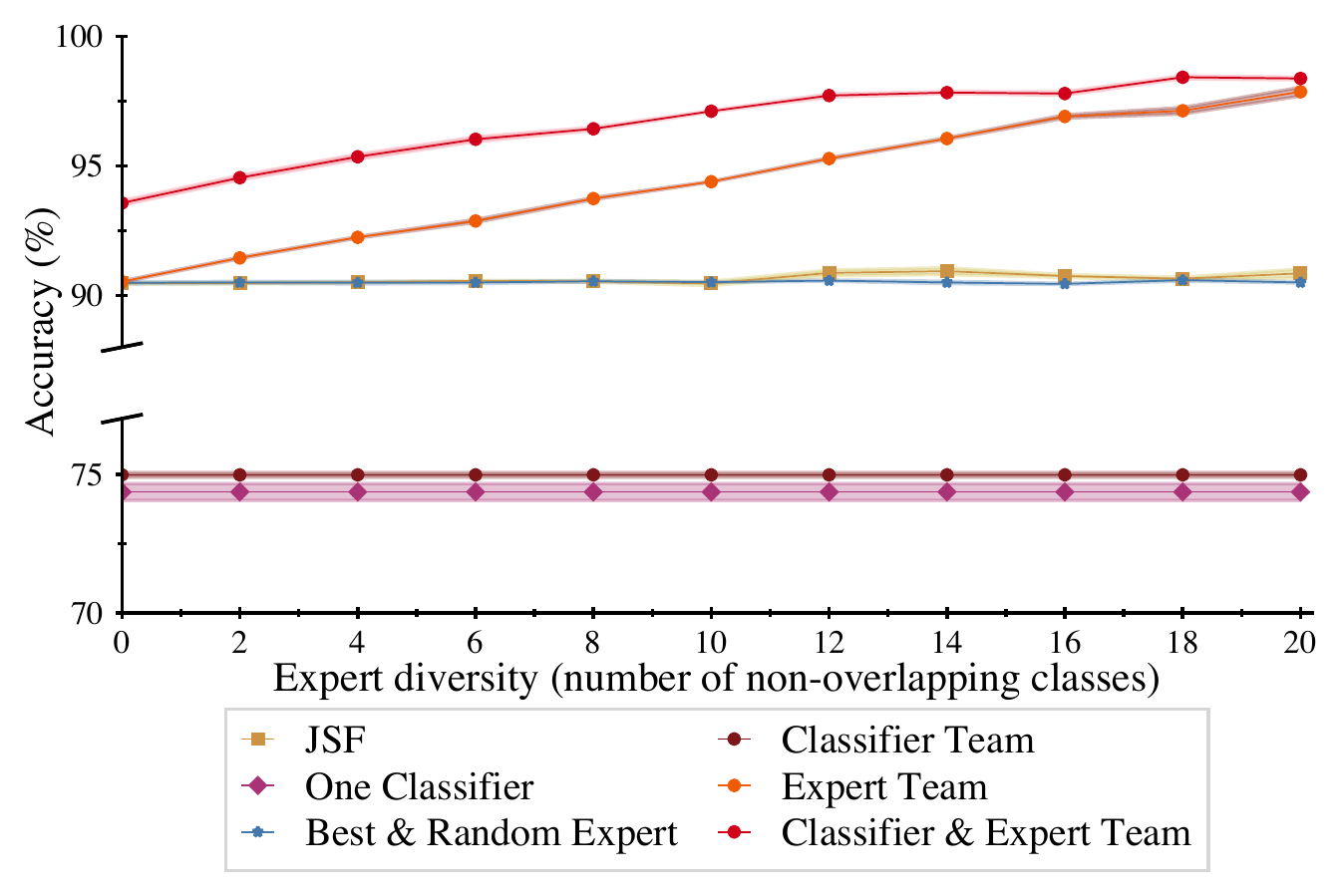}
    \caption{
    Team accuracy of our approach and the baselines with increasing human expert diversity (number of non-overlapping classes) on CIFAR-100. Shaded regions display standard errors.}
    \label{fig:expert_diversity}
\end{figure}
\begin{figure*}[h]
    \centering
    \subfloat[\centering Accuracy of human experts]{{\includegraphics[width=4.3cm, height=2.6cm]{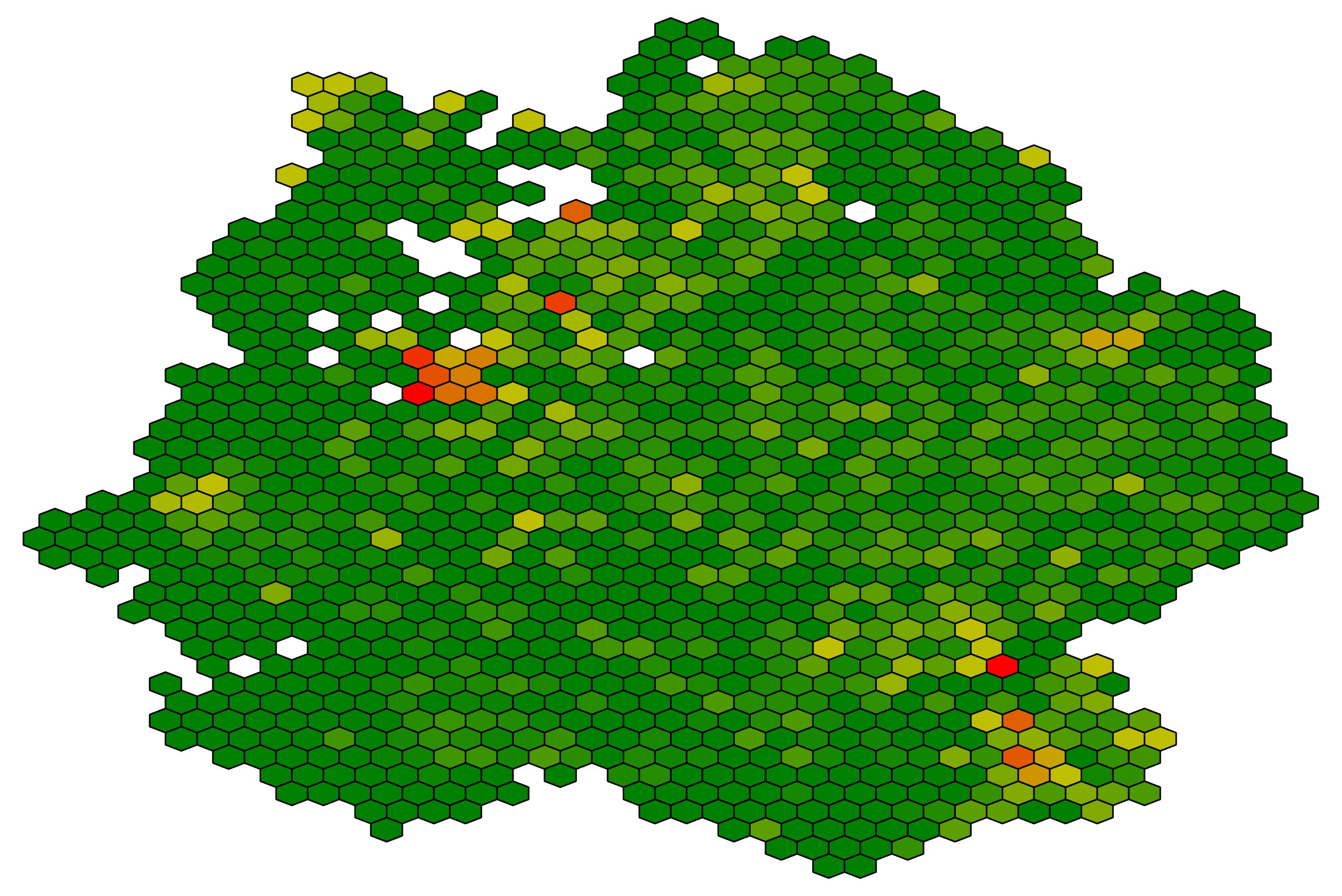}\label{fig:experts_accuracy}}}
    \subfloat[\centering Accuracy of classifier]{{\includegraphics[width=4.3cm,height=2.6cm]{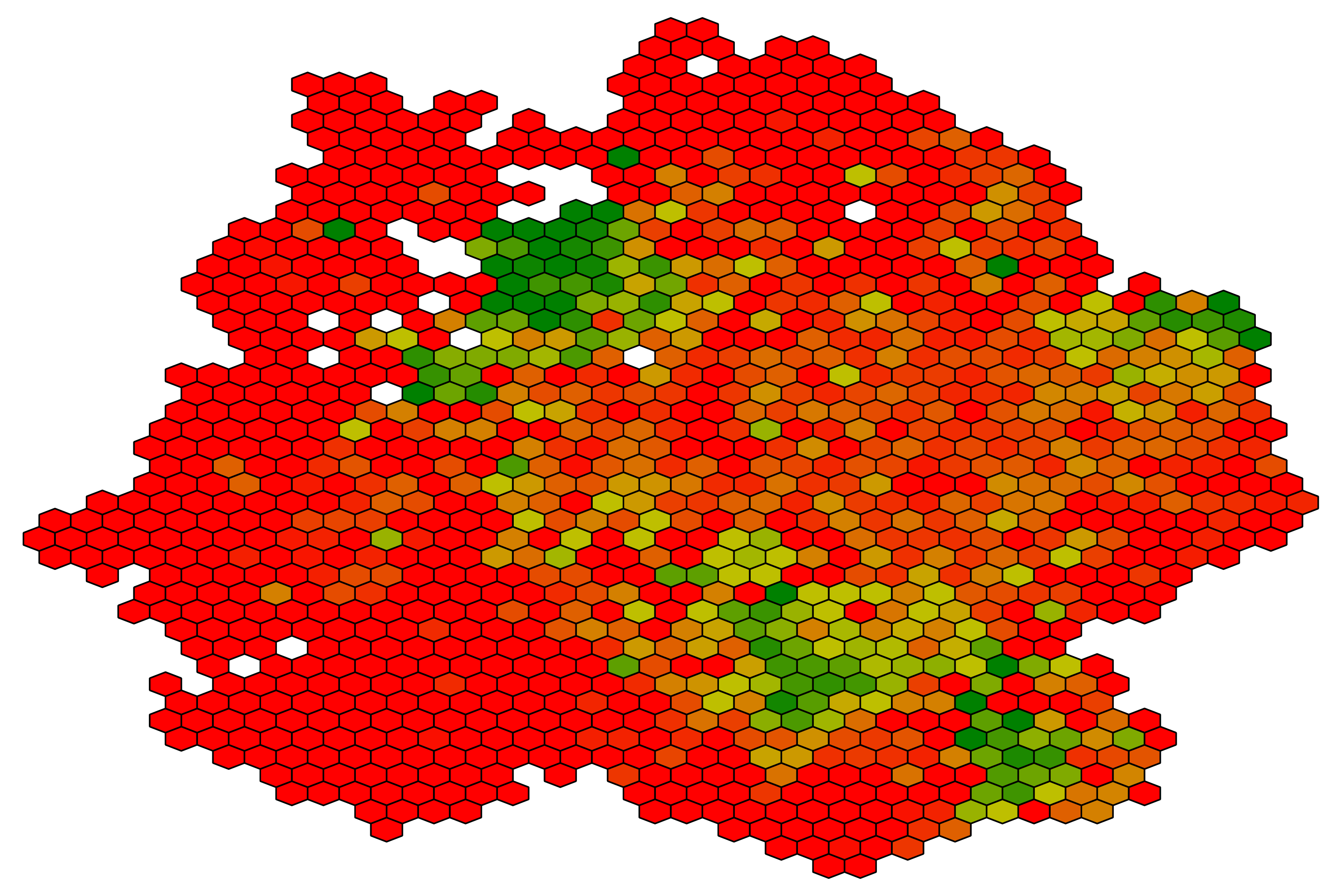}\label{fig:machine_accuracy}}}
    \includegraphics[width=1.2cm,height=2.6cm]{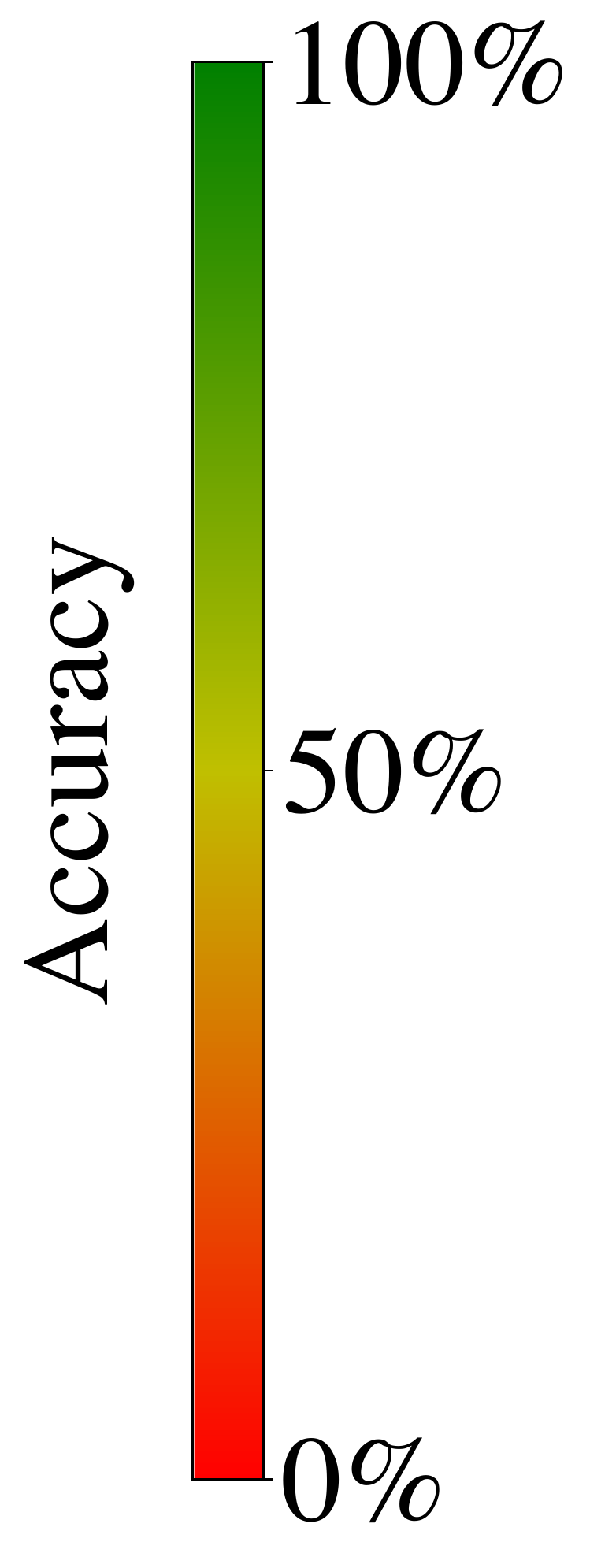}
    \qquad
    \subfloat[\centering Instance allocation]{{\includegraphics[width=4.3cm,height=2.6cm]{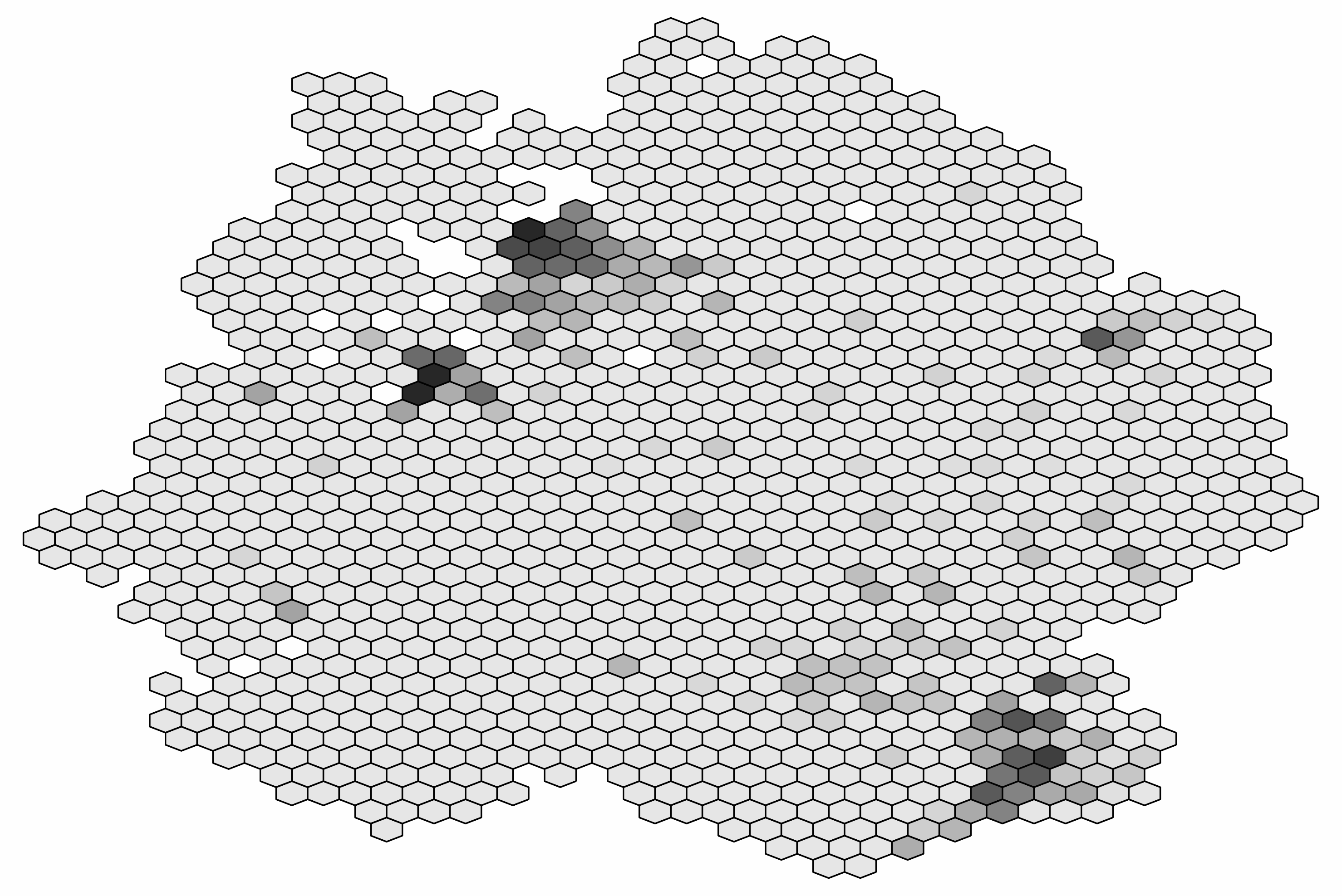}\label{fig:allocation}}}
    \includegraphics[width=1.2cm,height=2.6cm]{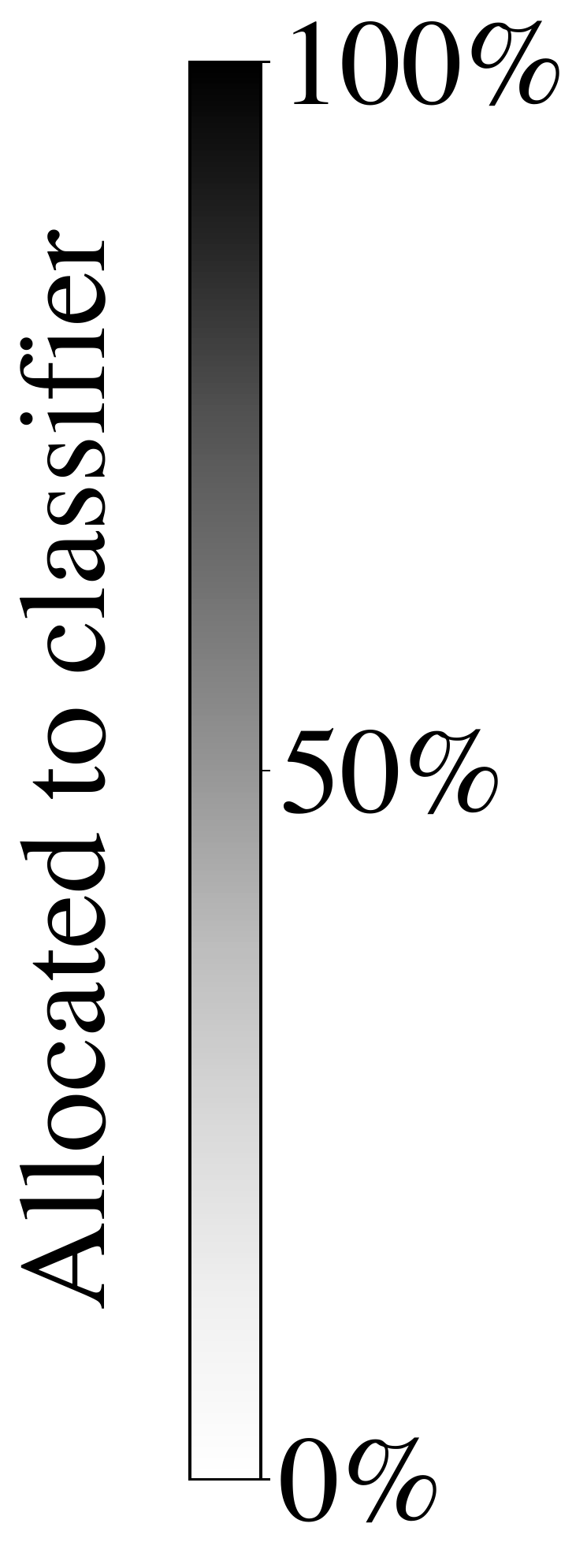}
    \caption{Visualization of the team members' performance---human experts' accuracy in (a) and classifier accuracy in (b)---and instance assignment by the allocation system in (c) on a subset of the CIFAR-100 input space.}
    \label{fig:qualitative_example} 
\end{figure*}

\paragraph{Dataset and Expert Generation.} The dataset consists of 60,000 images drawn from 100 subclasses which are grouped into 20 superclasses of equal size. Using the 20 superclasses for the classification task and the 100 subclasses for synthetic human expert generation allows us to model their competencies with more granularity compared to the Twitter dataset. For the first analysis, we simulate human experts in the following way. By design, for each of \(m\) human experts in the team, we draw \(k_j \sim \mathcal{N}(70,5) \,\, \forall j \in \{1, \dots, m\}\) from a normal distribution with \(\lfloor k_j \rfloor\) denoting the number of subclasses that each human expert predicts perfectly. For each of \(m\) human experts, the set of perfectly predicted subclasses is sampled uniformly at random without replacement from all 100 subclasses and mapped to the respective superclasses. For the remaining subclasses, each expert predicts uniformly at random across all superclasses. For the second analysis, we select \(m=2\) human experts over \(i \in \mathbb{N} \) runs. To isolate the influence of diversity, we vary the capabilities of the second human expert per run while keeping the ones of the first fixed. For \(i = 1\), we select the first 90 subclasses \(\{1,\ldots,90\}\) in alphabetical order for which both human experts predict the corresponding superclass perfectly. For the remaining subclasses, both experts predict uniformly at random across all superclasses. We then vary the competencies of the second human expert such that for run \(i\) the perfectly predicted subclasses are \(\{i,\ldots, 89+i\}\). By increasing \(i\) per run by 1, the number of subclasses in which only one of both human experts predicts perfectly increases by 2 subclasses resulting in a diversity increase. For the third analysis, we resort to one of these diversity scenarios.

\paragraph{Model.} We use a ResNet-18 model \cite{he2016deep} pretrained on ImageNet 
as a fixed feature extractor with 512 output units. These features are used to train our approach and the algorithmic baselines. The classifier and allocation system are modeled by a neural network consisting of a single hidden layer with 100 units followed by a ReLU activation.

\paragraph{Experimental Setup.} For the first analysis, we assess the performance of teams with 2 to 10 human experts with a normally distributed number of perfectly predicted subclasses as previously described. For the second analysis, we vary \(i\) from 1 to 11 for the second human expert to gradually increase the number of non-overlapping subclasses. As a result, diversity is
0 for run \(i=1\) and 20 for run \(i=11\). For the third analysis, we select one of the diversity scenarios---namely where the first human expert is defined as \(\{1,\ldots,90\}\) and the second one as \(\{6,\ldots,95\}\)---to visualize how instances are distributed among classifier and human experts relative to each team member's respective competencies.

We divide the 50,000 training images into a training and validation split of \(80 \%\) and \(20 \%\) and use the 10,000 test images as the test split. We train the models for 100 epochs using Adam optimizer with a learning rate of \(5 \cdot 10^{-3}\), a cosine annealing learning rate scheduler, weight decay of \(5 \cdot 10^{-4}\), and a batch size of 512. We apply early stopping on the validation split and report the results on the test split. We repeat the experiments 5 times with different seeds.

\paragraph{Results.} First, consistent with the results from \Cref{sec:section_hate_speech}, we find that our approach benefits from an increasing number of human experts. This is in line with our expectations as the probability that at least one expert has strengths in distinct subclasses or on all subclasses of a superclass is higher with an increasing number of human experts. In both cases, the classifier can focus on an even smaller area of the input space. On average, our approach (\textit{Classifier \& Expert Team}) outperforms the \textit{JSF} baseline by 19.77\%, the \textit{Best Expert} baseline by 19.73\%, the \textit{One Classifier} baseline by 25.86\%, the \textit{Random Expert} baseline by 30.75\%, and the \textit{Classifier Team} by 24.88\%. Even though the contribution from the classifier in our approach diminishes the more human experts become part of the team, it leverages each individual team member's capabilities best. This can be observed from a performance gain of 0.98\% compared to the \textit{Expert Team}. We refer to Figure A1 in Appendix A for an additional visualization. 

Second, we find that for the \textit{Classifier \& Expert Team}, a higher diversity of human team members' capabilities contributes to higher team performance as displayed in \Cref{fig:expert_diversity}. Even though the contribution of the classifier diminishes in the presence of increasingly diverse human experts, it adds an average performance gain of 2.40\% across all runs compared to the \textit{Expert Team}. Contrary, the \textit{JSF} baseline does not benefit from a diversity increase but instead assigns the instances to the best human expert. Note that we aggregate the \textit{Best Expert} and \textit{Random Expert} baseline in \Cref{fig:expert_diversity} as both human experts have the same accuracy by design.

Finally, we illustrate the capabilities of our approach in \Cref{fig:qualitative_example}: (1) the classifier's compensation for human experts' weaknesses and (2) the allocation system's assignment of instances relative to each team member's competencies. For the visualization, we reduce the hidden layer's features to 2 dimensions using the UMAP algorithm \cite{mcinnes2018umap} and aggregate a set of images in each hexagon. \Cref{fig:experts_accuracy} and \Cref{fig:machine_accuracy} show the accuracy of 2 human experts and the classifier, respectively. In combination, the illustrations reveal that the classifier primarily learns to optimize its performance on those images (green hexagons in \Cref{fig:machine_accuracy}) where human experts tend to be less accurate (yellow and red hexagons in \Cref{fig:experts_accuracy}). \Cref{fig:allocation} displays that the allocation system assigns those instances to the classifier where the human experts are less accurate. Here, dark areas indicate the assignment of the images to the classifier and brighter areas denote the allocation to one of the human experts.

\subsection{NIH Dataset}

Lastly, we evaluate our approach on a real-world chest X-ray dataset collected by the NIH Clinical Center \cite{NIHDataset,chestxray8}. In the previous experiments, we induced a synthetic connection between certain features (i.e., the tweet dialect or the image subclass) and the human experts' capabilities. Contrary, for the NIH dataset, it remains ex-ante unclear to which degree features are associated with the human experts' competencies.
\begin{table}[t]
    \small
    \centering
    \resizebox{\linewidth}{!}{%
    \begin{tabular}{lllll}\toprule
         & 
         \multicolumn{4}{c}{Team Accuracy (\%)}
         
         \\\cmidrule{2-5}

        Method & 
        \multicolumn{1}{c}{[\(F\), 357, 117]} &
        \multicolumn{1}{c}{[\(F\), 357, 121]} & 
        \multicolumn{1}{c}{[\(F\), 249, 124]} & 
        \multicolumn{1}{c}{[\(F\), 249, 296]} 
        \\\midrule
        
        \textbf{Baselines:} \\
        
        - JSF &
        94.69 ($\pm$ 0.25) &
        90.14 ($\pm$ 0.17) & 
        88.69 ($\pm$ 0.32) &
        90.92 ($\pm$ 0.13) 
        \\
        
        - One Classifier & 
        83.13 ($\pm$ 0.21) &
        84.70 ($\pm$ 0.13) & 
        84.59 ($\pm$ 0.12) &
        83.63 ($\pm$ 0.08) 
        \\

        - Random Expert & 
        89.52 ($\pm$ 0.29) &
        89.05 ($\pm$ 0.19) & 
        88.45 ($\pm$ 0.19) &
        87.20 ($\pm$ 0.36) 
        \\
         
        - Best Expert &
        95.34 ($\pm$ n/a)~~ &
        91.01 ($\pm$ n/a) & 
        88.84 ($\pm$ n/a)~~ &
        91.31 ($\pm$ n/a)~~ 
        \\

        - Classifier Team & 
        83.44 ($\pm$ 0.16) & 
        85.09 ($\pm$ 0.17) & 
        84.83 ($\pm$ 0.23) & 
        83.69 ($\pm$ 0.27) 
        \\
        
        - Expert Team & 
        95.34 ($\pm$ 0.04) & 
        91.08 ($\pm$ 0.04) & 
        88.33 ($\pm$ 0.11) & 
        91.29 ($\pm$ 0.11)  
        \\
        
        \textbf{Our Approach:} \\
        
         - C. \& E. Team & 
        \textbf{95.45} ($\pm$ 0.06)  & 
        \textbf{91.72} ($\pm$ 0.12)  & 
        \textbf{90.36} ($\pm$ 0.15)  & 
        \textbf{91.51} ($\pm$ 0.13) 
        \\

         \bottomrule
    \end{tabular}
    }
 \caption{Team accuracies of our approach and the baselines including standard errors for 4 different human-AI teams consisting of the classifier \(F\) and two radiologists on the NIH dataset.}
    \label{tab:results_nih}
\end{table}

\paragraph{Dataset.} The dataset comprises radiologists' annotations for 4 radiographic findings on 4,374 chest X-ray images from the ChestX-ray8 dataset \cite{NIHDataset,chestxray8}. We focus on the occurrence of the clinically important finding \textit{airspace opacity} which has a prevalence (i.e., the percentage of patients in the dataset that is affected with this disease) of 49.50\%. For each X-ray image, the ground-truth label was adjudicated by a panel of 3 radiologists serving as the ``gold standard'' label. Additionally, the annotations of each of the three radiologists from a cohort of 22 human experts are reported per image. Each radiologist can be identified by a unique ID of which we list the last three digits due to space constraints.

\paragraph{Model.} We use a ResNet-18 model pretrained on the  CheXpert dataset---a different X-ray dataset for chest radiograph interpretation \cite{Chexpert}---which serves as a fixed feature extractor with 512 output units. For our approach and the algorithmic baselines, we use these features for model training. The classifier and allocation system are modeled by a single hidden layer neural network with 30 units and a ReLU activation function.

\paragraph{Experimental Setup.} We conduct experiments with 4 pairs of radiologists that share more than 800 images labeled by both radiologists. Due to insufficient intersection, a third radiologist could not be included in the team. We perform a 10-fold cross-validation while ensuring that images from one patient are not spread across different folds. To keep the per fold performance of the radiologists close to their overall performance, we use stratification. For each cross-validation run, we use 7 folds for training and 2 folds for validation to perform early stopping as the team performance on one validation fold is not always a good proxy for the team performance on the test fold due to the small dataset size. The remaining fold serves as the test data. We train the models for 20 epochs using Adam optimizer with a learning rate of \(1 \cdot 10^{-3}\), weight decay of \(5 \cdot 10^{-4}\), and a batch size of 64. We repeat the experiments 5 times with different seeds.
\begin{table}[t]
    \small
    \centering
    \resizebox{\linewidth}{!}{%
    \begin{tabular}{lccc}\toprule
         & 
         &
        \multicolumn{1}{c}{Individual Accuracy (\%)} &
         \\\cmidrule{2-4}
        
        Instances allocated to & 
        Classifier \(F\) & 
        ID = 357 & 
        ID = 117 \\\midrule
         
        Classifier \(F\) & 
        \textbf{97.73} ($\pm$ 1.76) &
        78.93 ($\pm$ 2.47) &
        96.43 ($\pm$ 1.86) \\
        
        ID = 357 &
        31.88 ($\pm$ 4.73) &
        \textbf{98.78} ($\pm$ 0.32) &
        98.55 ($\pm$ 0.34) \\
        
        ID = 117 & 
        61.69 ($\pm$ 1.15) &
        79.90 ($\pm$ 0.29) &
        \textbf{94.67} ($\pm$ 0.02) \\
        
         \bottomrule
    \end{tabular}
    }
 \caption{Individual accuracy of the classifier, radiologist ID = 357, and radiologist ID = 117 on the subset of instances assigned to each of them. Additionally, the team members' performances on the subsets not assigned to them are displayed.}
 \label{tab:nih_qualitative}
\end{table}

\paragraph{Results.} \Cref{tab:results_nih} compares the performance of our approach with the baselines. In detail, the performance of our approach (\textit{Classifier \& Expert Team}) outperforms the other baselines. Particularly, in high-stakes decision-making (e.g., medicine), the performance improvements can contribute to further reducing potentially severe consequences of misdiagnoses. A closer look at the subset of images assigned to each human expert and the classifier shows that the performance of each team member is highest on the subset assigned to them. \Cref{tab:nih_qualitative} illustrates this allocation by the diagonal values using the example of the human-AI team [\(F\), 357, 117]. Thus, the classifier learns and is assigned to an area of the input space where it is more accurate than the radiologists. Furthermore, the allocation system assigns the remaining instances between the two radiologists based on their competencies.

\section{Conclusion}
In this work, we leverage the complementary capabilities of human experts and AI models by jointly training a classifier together with an allocation system. While the classifier augments human experts' weaknesses, the allocation system assigns instances to either one of the human experts or the classifier based on their individual capabilities. We provide experimental evidence with synthetic and real human expert annotations for the ability of such teams to not only outperform prior work but also to achieve superior performance results none of the team members could have accomplished alone. Furthermore, we show that our approach can cope with different degrees of human expert diversity and generally benefits from a higher number of human team members. As we do not constrain the availability and capacity of experts at run time, opportunities for future research could explore load balancing strategies or alternative assignment strategies if the most accurate team member is not available. Lastly, we assume access to the human experts' predictions for all training data, implying a significant manual labeling effort. To address this limitation, further work could extend our approach to leverage data that is not annotated by all human experts in a team.

\section*{Ethical Statement}

As this work pursues the training of an algorithm that augments the weaknesses of human experts, it is subject to important ethical considerations. Our goal is to realize improved performance compared to full automation and sole human effort which can be beneficial in high-stakes decision-making. However, the approach could theoretically be used to estimate the contribution of individual team members to the team's overall performance. This could be used to exclude insufficiently performing experts from the team. However, identifying the weaknesses of individual experts also provides an opportunity to educate them through targeted training.

\bibliographystyle{named}
\bibliography{ijcai22}

\end{document}